\documentclass[11pt]{article}
\usepackage[margin=0.9in]{geometry}
\usepackage{times}
\usepackage[numbers,sort&compress]{natbib}

\usepackage{amsmath,amsfonts,bm}

\def\eqref#1{equation~\ref{#1}}

\def\1{\bm{1}}

\DeclareMathAlphabet{\mathsfit}{\encodingdefault}{\sfdefault}{m}{sl}
\SetMathAlphabet{\mathsfit}{bold}{\encodingdefault}{\sfdefault}{bx}{n}

\usepackage{graphicx}
\usepackage{booktabs}
\usepackage{hyperref}
\usepackage{url}
\usepackage{xcolor}

\title{MAMHOI: Factorizing Scene-Aware Human-Object Interaction through Affordances}

\author{
Mingyuan Lei \qquad Yoonchang Sung \qquad Tat-Jen Cham\\[0.5em]
\normalsize College of Computing and Data Science\\
\normalsize Nanyang Technological University, Singapore
}
\date{}

\begin{document}

\maketitle

\begin{abstract}
Generating realistic human-object interactions (HOI) in complex 3D scenes requires two complementary capabilities: reasoning about interaction feasibility in the environment and synthesizing realistic human-object motion. However, supervision for these capabilities is rarely available jointly at scale. Human-scene datasets provide rich information about environment-aware motion, while human-object datasets capture detailed interaction dynamics, yet paired human-object-scene data remain scarce. We present MAMHOI, an affordance-mediated factorization for scene-aware human-object interaction generation. MAMHOI factorizes scene-aware HOI generation through an explicit motion-affordance interface between scene understanding and motion synthesis: a scene-conditioned model first predicts where and how an interaction can be feasibly executed, and an affordance-conditioned HOI model then generates the corresponding human-object motion. This factorization allows scene understanding and interaction dynamics to be learned from complementary sources of supervision without requiring paired human-object-scene data. Experiments in complex indoor environments show that MAMHOI reduces object--scene penetration while better preserving human--object interaction quality, yielding more realistic and physically feasible scene-aware interactions.
\par\smallskip
\noindent\textbf{Project page:} \url{https://leimingyuan.github.io/MAMHOI-project-page/}
\end{abstract}

\section{Introduction}

Generating realistic human-object interactions (HOIs) in complex 3D environments requires more than synthesizing plausible human and object motion. The generated interaction must also adapt to the surrounding scene: the human must approach and manipulate the object without violating scene geometry, while the object must move through feasible regions and reach its intended target. Recent generative motion models have substantially improved the realism, diversity, and controllability of human motion and human-object interaction synthesis \citep{mdm,mofusion,mld,t2mgpt,motiongpt,momask}. However, many of these models are developed in scene-free or simplified environments and therefore do not directly address how an interaction should adapt to complex scene geometry.
\begin{figure*}[t]
    \centering
    \includegraphics[width=\linewidth]{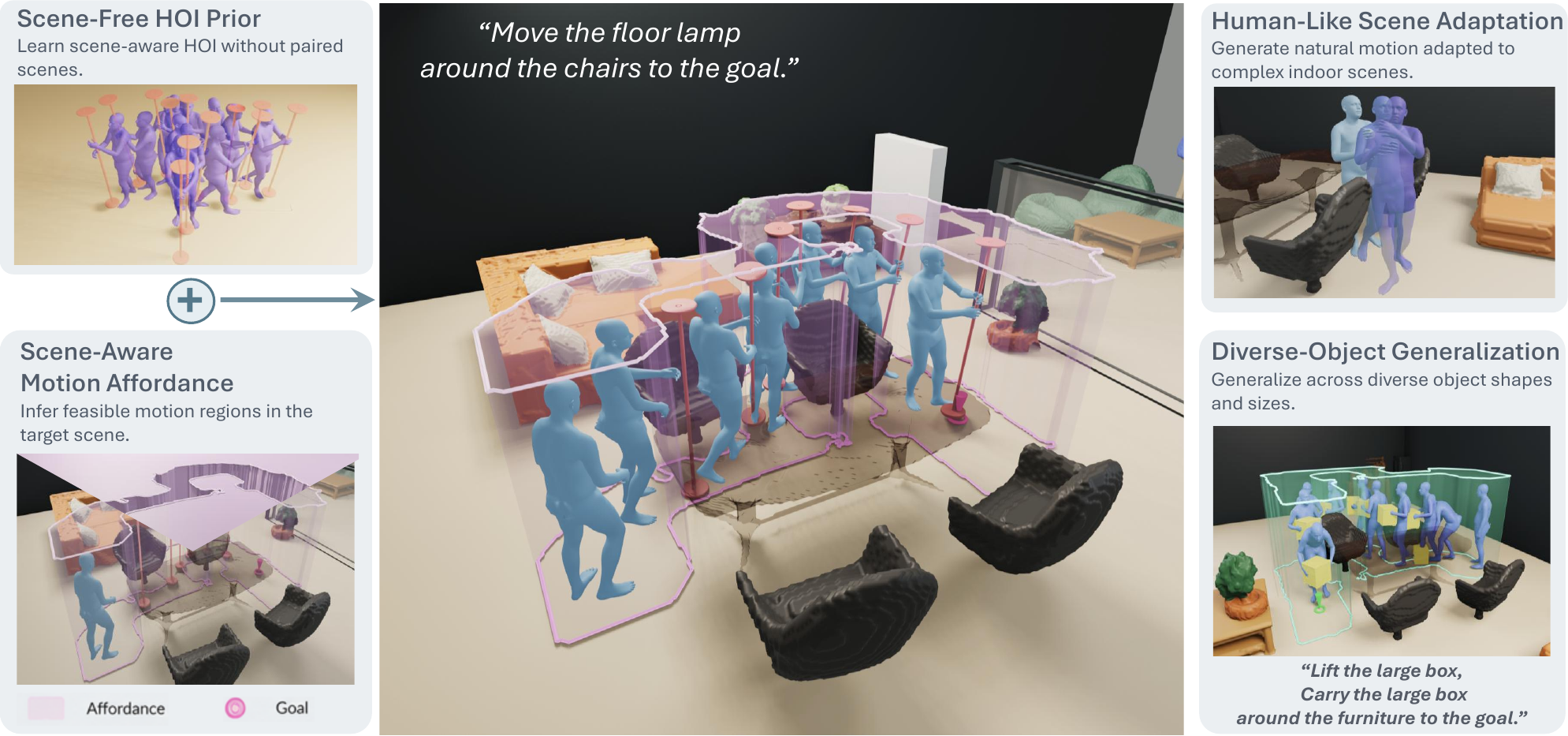}
    \caption{MAMHOI learns from Scene-Free HOI priors and transfers the ability to complex environments through scene-aware motion affordances. This factorization enables physically plausible interactions across cluttered scenes and diverse objects.}
    \label{fig:teaser}
\end{figure*}
Scene-aware HOI generation requires two complementary capabilities. First, the model must reason about \emph{interaction feasibility}: where and how an interaction can be executed given the scene layout, object location, and task objective. Second, it must model \emph{interaction dynamics}: how the human and object should move together while maintaining realistic motion, contact, and manipulation behavior. Although these capabilities must ultimately operate together, supervision for them is rarely available jointly at scale. Human-scene datasets provide rich examples of environment-aware behavior \citep{humanise,trumans,lingo}, while human-object datasets capture detailed synchronized interaction dynamics \citep{chois,interdiff}. In contrast, collecting human-object-scene interactions that jointly cover diverse motions, objects, and scene geometries is substantially more difficult.

Existing methods bridge this gap in several ways. Some methods incorporate scene constraints through object waypoints, interaction anchors, or trajectory refinement \citep{chois,hosig}, while others construct synthetic scene-interaction pairs and directly train scene-conditioned generators \citep{unihm}. Recent work further shows that human-object and human-scene data can be combined through hybrid training without requiring fully annotated human-object-scene interactions \citep{infbagel}. These approaches demonstrate the value of leveraging complementary data sources, but leave a fundamental modeling question open: \emph{how should scene-level interaction feasibility and detailed human-object dynamics be connected when their strongest supervision comes from different sources with distribution gap?}

We address this question by explicitly factorizing scene-aware HOI generation through an intermediate \emph{motion affordance}. Rather than directly learning a single distribution over human motion, object manipulation, and scene geometry, Motion-Affordance-Mediated Human--Object Interaction (MAMHOI) separates scene understanding from human-object motion synthesis. The scene-understanding component predicts an affordance representation that captures task-relevant regions in which the interaction can be feasibly executed, while the motion generator synthesizes synchronized human-object motion conditioned on this representation. The affordance therefore serves as an explicit interface between the two capabilities: it communicates the scene-dependent constraints needed for interaction without prescribing the final human-object trajectory.

Concretely, MAMHOI predicts motion affordance from scene geometry, object-motion cues, and language instructions, and converts the predicted affordance into local scene occupancy voxel grids around the human, object, and target region. These representations condition a diffusion-based human-object motion generator, allowing detailed interaction dynamics learned from motion-rich HOI data to adapt to complex environments. Planning is used only to provide coarse object-level structural cues rather than to determine the resulting human motion. This factorization allows scene reasoning and interaction synthesis to exploit complementary sources of supervision without requiring densely paired human-object-scene data.

Experiments across interaction-only and scene-aware settings demonstrate the benefit of this formulation. MAMHOI preserves high-quality HOI dynamics while substantially improving scene compatibility, reducing penetrations without compromising object trajectory adherence.

Our main contributions are summarized as follows:
\begin{itemize}
    \item We introduce \textbf{MAMHOI}, an affordance-mediated factorization of scene-aware HOI generation that separates scene-level interaction feasibility from detailed human-object motion synthesis, allowing the two capabilities to exploit complementary sources of supervision.
    \item We formulate motion affordance as an explicit interface between scene reasoning and interaction synthesis, conveying task-relevant environmental constraints to the motion generator without restricting generation to a single prescribed trajectory.
    \item We demonstrate that MAMHOI improves scene feasibility, particularly by reducing object--scene penetration, while preserving high-quality human--object interaction dynamics across complex indoor environments.
\end{itemize}
\section{Related Work}
\subsection{Human-Centric Motion Generation}
Text-conditioned human motion generation aims to synthesize realistic motion from natural-language descriptions. Early methods use VAE- or GAN-based formulations, while recent diffusion and autoregressive models improve motion quality, diversity, and semantic alignment~\citep{humanml3d,temos,action2motion,mdm,mofusion,motiondiffuse,mld,t2mgpt,motiongpt,momask}. HOI generation extends this task by modeling human motion jointly with object manipulation, contact, and relative human-object movement~\citep{interdiff,goal,imos,physhoi}. Recent HOI methods incorporate explicit contact constraints, object affordances, relation-aware denoising, object geometry, and trajectory conditions to generate coordinated human-object motion~\citep{cghoi,hoidiff,thor,interdreamer}. In parallel, human-scene interaction (HSI) methods condition human motion on static scene geometry represented by point clouds, meshes, voxels, or egocentric features~\citep{prox,humanise,trumans,lingo}. 
Afford-Motion~\citep{wang2024move} further introduces scene affordance as an intermediate representation for language-guided HSI, defining it as a continuous distance field between human joints and 3D scene surfaces.
While sharing the idea of affordance-mediated generation, our motion affordance differs in both representation and role: it encodes the feasible spatial support of the joint human--object interaction rather than body--scene proximity, and serves to transfer scene-level feasibility to an HOI generator learned from scene-free human--object motion. Physics-based simulation and reinforcement learning have also been used to produce physically feasible interactions with static scene elements~\citep{scenediff,unihsi}. HOI methods generally emphasize manipulation in scene-free or simplified environments, whereas HSI methods primarily model human navigation and interaction with static scene elements.
\subsection{Scene-Aware Human-Object Interaction Generation}
Recent methods have begun to address scene-aware human-object interaction generation. CHOIS~\citep{chois} generates synchronized human-object motion from language, initial states, object geometry, and sparse object waypoints that can be provided by a high-level scene planner. Its scene constraints are therefore conveyed mainly through externally planned waypoints rather than an explicit representation of feasible interaction regions. HOSIG~\citep{hosig} decomposes the task into scene-aware grasp-pose generation, heuristic navigation, and scene-guided motion synthesis. However, both its interaction anchors and controllable motion generator rely on paired human-object-scene interaction data, and its released setup demonstrates only limited object coverage.

Another line of work addresses the scarcity of paired interaction-scene data through data augmentation. UniHM~\citep{unihm} constructs synthetic scene-motion pairs by placing existing motions into collision-free regions of indoor scenes and trains a unified scene-conditioned generator on the resulting data. Such augmentation primarily encourages the model to identify feasible regions in which a motion can be executed, while its waypoint conditioning does not guarantee exact object trajectory adherence. InfBaGel~\citep{infbagel} further combines synthesized human-object-scene interactions with real human-scene interaction data through hybrid training. While this improves scene feasibility, jointly learning from heterogeneous motion distributions introduces a trade-off between scene-level physical feasibility and preserving task-specific human-object interaction priors.

Our method instead avoids constructing or jointly fitting heterogeneous paired human-object-scene data. We represent scene constraints through motion affordance and learn scene feasibility separately from human-object motion, allowing scene knowledge to guide where interactions are feasible while preserving the interaction distribution learned from dedicated human-object data.

\section{Methodology}
\begin{figure*}[t]
    \centering
    \includegraphics[width=1.0\linewidth]{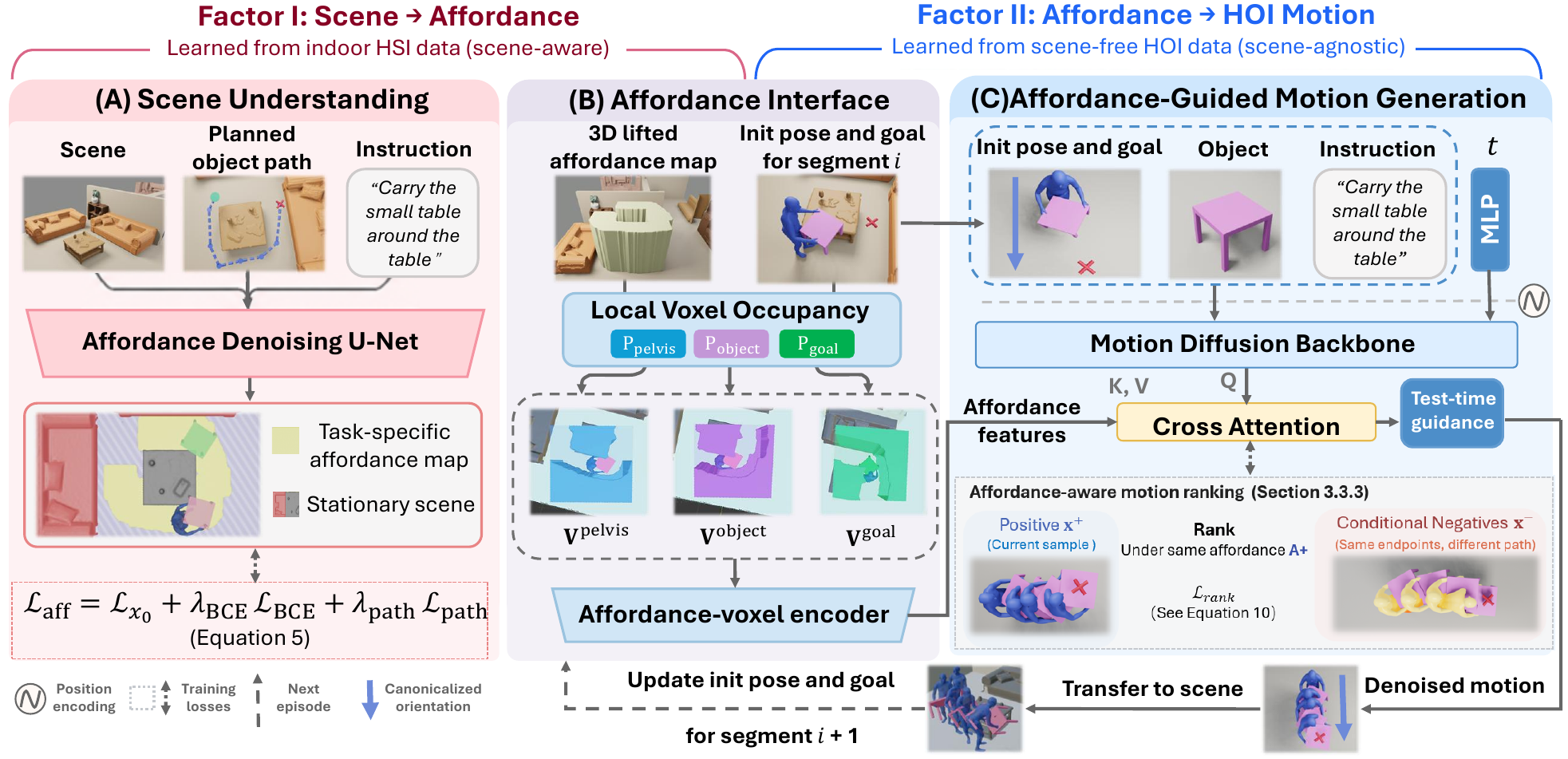}
\caption{
Overview of MAMHOI.
\textbf{(A)} The scene-understanding module predicts a task-specific affordance from the scene, planned object path, and instruction, where the instruction guides the task-relevant human placement along the object path.
\textbf{(B)} The affordance interface lifts the predicted affordance to 3D and queries local occupancy around the pelvis, object, and goal position of current segment, and encodes them into affordance features through a ViT-based Affordance-voxel encoder. 
\textbf{(C)} The resulting affordance features condition segment-wise HOI generation through cross-attention, with each generated segment initializing the next.
Through this design, scene-aware HOI generation is factorized into a \textbf{scene-to-affordance factor} learned from indoor HSI data and an \textbf{affordance-to-motion factor} learned from scene-free HOI data.
}
    \label{fig:Framework}
\end{figure*}
\subsection{Problem Formulation}

Given a 3D scene, a language instruction, an initial human--object state,
object geometry, and a planned object path, our goal is to generate a
synchronized human--object motion sequence $\mathbf{x}$ that is consistent
with both the task and the surrounding scene. Let $c_{\mathrm{task}}$ and
$c_{\mathrm{scene}}$ denote the task and scene conditions, respectively. The
general objective is
\begin{equation}
p(\mathbf{x}\mid c_{\mathrm{task}},c_{\mathrm{scene}}).
\end{equation}

We factorize this distribution through a sequence-level motion affordance $\mathbf{A}$ that summarizes the scene-dependent spatial constraints relevant to the interaction. We assume that, once this affordance is given, the detailed scene condition becomes approximately redundant for human--object motion synthesis. Under the conditional-independence assumptions detailed in Appendix~\ref{app:factored-objective},
\begin{equation}
\label{eq:factored-objective}
p(\mathbf{x}\mid c_{\mathrm{task}},c_{\mathrm{scene}})
\approx
\int
p(\mathbf{x}\mid \mathbf{A},c_{\mathrm{task}})
\,
p(\mathbf{A}\mid c_{\mathrm{task}}',c_{\mathrm{scene}})
\,d\mathbf{A}.
\end{equation}

Here, $c_{\mathrm{task}}'$ retains only the information required for affordance prediction. In our implementation, $c_{\mathrm{task}}'=(\mathbf{P},\mathbf{Text})$, where $\mathbf{P}$ denotes the planner-provided object path, while  $c_{\mathrm{task}}$ additionally contains object geometry and initial human--object states. This factorization separates the supervision required for scene reasoning from that required for detailed interaction synthesis, making it possible to learn them independently from HSI and HOI data. The predicted affordance then serves as the interface through which scene-dependent constraints are transferred to the motion generator.

\subsection{Data Representation}

We represent each interaction segment using human motion $\mathbf{H}$, object motion $\mathbf{O}$, object geometry $\mathbf{G}$, sparse motion conditions $\mathbf{C}_{\mathrm{motion}}$, and a scene-aware affordance condition $\mathbf{a}$. For $N_J$ human joints, human motion is represented by 3D joint positions and continuous 6D joint rotations, while object motion consists of its centroid and rotation. We denote the complete human--object motion sequence as $\mathbf{x}_0=[\mathbf{H},\mathbf{O}]$. Both human and object motion are transformed into the same canonical human-local coordinate frame to remove global orientation ambiguity and align the motion with the scene representation. Following CHOIS~\citep{chois}, object geometry is encoded using a Basis Point Set (BPS) descriptor $\mathbf{G}\in\mathbb{R}^{1024\times3}$. The masked condition $\mathbf{C}_{\mathrm{motion}}$ contains the initial human--object state, sparse object waypoints, and the object goal state.

We distinguish between the sequence-level 2D affordance map $\mathbf{A}$ predicted by the scene understanding model and the local 3D affordance condition $\mathbf{a}$ consumed by the motion generator. Specifically, $\mathbf{a}$ consists of three local occupancy voxel grids,
$\mathbf{V}^{\mathrm{pelvis}}, \mathbf{V}^{\mathrm{obj}}, \mathbf{V}^{\mathrm{goal}}
\in \{0,1\}^{32\times32\times32}$,
centered at the human pelvis, current object position, and object goal, respectively. These voxels capture local spatial constraints relevant to the given state, task and goal, and share the same canonical orientation as the motion representation. We concatenate them along the channel dimension as
\begin{equation}
\mathbf{a} =
\mathbf{V}^{\mathrm{pelvis}}
\oplus
\mathbf{V}^{\mathrm{obj}}
\oplus
\mathbf{V}^{\mathrm{goal}}
\in
\{0,1\}^{3\times32\times32\times32},
\end{equation}
where $\oplus$ denotes channel-wise concatenation. The text instruction, object geometry, and affordance condition are encoded as
$\mathbf{e}_{\mathrm{text}}$,
$\mathbf{e}_{\mathrm{geo}}$, and
$\mathbf{e}_{\mathrm{aff}}$
using CLIP, an MLP, and a ViT-based Affordance-voxel encoder, respectively, and jointly condition the model.
\subsection{Motion-Affordance-Mediated Human--Object Interaction Framework}
\subsubsection{Scene Understanding Model}
\label{sec:Scene Understanding}
We define a motion affordance as the spatial support within which a
human--object interaction can feasibly evolve. It is represented as a sequence-level binary 2D region
covering the interaction space of both the human and the manipulated object.
Given a 2D scene depth map, a planner-provided object path, and a language
instruction, the Scene Understanding Model predicts this affordance for the
target interaction:
\begin{equation}
\hat{\mathbf{A}} =
g_{\psi}
(
\mathbf{S}_{\mathrm{depth}},
\mathbf{P},
\mathbf{e}_{\mathrm{text}}
),
\end{equation}
where $\mathbf{S}_{\mathrm{depth}}$ denotes the 2D scene depth map, $\mathbf{P}$ denotes the planner-provided object path, and $\mathbf{e}_{\mathrm{text}}$ is the CLIP text embedding.

The model first encodes the scene depth map using a residual 2D U-Net,
while a path encoder transforms $P$ into a sequence of positional
features and CLIP encodes the language instruction. At the U-Net
bottleneck, the spatial scene features serve as queries in a multi-head
cross-attention layer, while the path and text features provide keys and
values. The conditioned spatial features are then decoded into a 2D
affordance probability map, which is thresholded at $\tau$ to obtain the
binary affordance mask $\hat{A}$. This allows the model to identify
task-relevant regions by jointly reasoning about scene geometry,
object-level navigation intent, and language semantics.

During training, the scene understanding model is supervised using ground-truth motion affordances constructed from the corresponding scene-aware motion trajectories. We optimize
\begin{equation}
\mathcal{L}_{\mathrm{aff}}
=
\mathcal{L}_{x_0}
+
\lambda_{\mathrm{BCE}}\,\mathcal{L}_{\mathrm{BCE}}
+
\lambda_{\mathrm{path}}\,\mathcal{L}_{\mathrm{path}},
\end{equation}
where the auxiliary terms emphasize affordance prediction around the planner path and explicitly encourage path coverage, respectively. Detailed definitions and loss weights are provided in Appendix~\ref{app:affordance_training}.

We first lift the predicted 2D affordance into the scene coordinate system using the scene geometry, obtaining a 3D affordance representation $\mathcal{V}(\hat{\mathbf{A}},\mathbf{S}_{\mathrm{depth}})$. For each autoregressive segment, we then query this representation around three interaction-centric reference points:
\begin{equation}
\mathbf{V}^{k} =
\mathrm{Query}
\left(
\mathcal{V}(\hat{\mathbf{A}},\mathbf{S}_{\mathrm{depth}}),
\mathbf{c}^{k}
\right),
\qquad
k \in
\{\mathrm{pelvis},\mathrm{obj},\mathrm{goal}\}.
\end{equation}
Each queried volume has size $32\times32\times32$. The three local affordance voxels are concatenated to form $\mathbf{a}$ and passed to the motion diffusion backbone.

\subsubsection{Affordance-Conditioned Motion Diffusion Backbone}
During training on scene-free HOI data, we construct the affordance condition from the observed human–object motion as discussed in \ref{exp:evaluationset}. At inference, the condition is supplied by the scene-understanding model. 
Following CHOIS~\citep{chois} and the conditional denoising formulation used
in InfBaGel~\citep{infbagel}, we adopt a similar backbone for synchronized
human--object motion synthesis.
Our main architectural difference lies in how scene information is incorporated:
the predicted local affordance voxels are encoded by a ViT-based scene encoder
and injected into the motion backbone through multi-head cross-attention.
This allows scene information to condition the motion generator through the
learned affordance representation rather than being directly coupled to the
human--object motion representation.

For each autoregressive segment, let
$\mathbf{x}_0 \in \mathbb{R}^{T\times D_x}$ denote the clean $T$-frame
human--object motion segment and $\mathbf{x}_n$ its noisy version at diffusion
step $n$. Following the clean-sample prediction strategy, the denoising network
predicts $\hat{\mathbf{x}}_0$ from the noisy motion and multimodal conditions:
\begin{equation}
\hat{\mathbf{x}}_0 =
f_{\theta}
(
\mathbf{x}_n,
n,
\mathbf{C}_{\mathrm{motion}},
\mathbf{e}_{\mathrm{text}},
\mathbf{e}_{\mathrm{aff}},
\mathbf{e}_{\mathrm{geo}}
).
\end{equation}
Here, $\mathbf{C}_{\mathrm{motion}}$ denotes the masked sparse motion conditions
defined above, while $\mathbf{e}_{\mathrm{text}}$, $\mathbf{e}_{\mathrm{aff}}$,
and $\mathbf{e}_{\mathrm{geo}}$ denote the text, affordance, and object-geometry
features, respectively.

The local affordance voxels are encoded by a ViT-based scene encoder:
\begin{equation}
\mathbf{e}_{\mathrm{aff}} =
E_{\mathrm{ViT}}(\mathbf{a}).
\end{equation}
The resulting affordance features are used as keys and values in a multi-head
cross-attention layer, while intermediate motion features from the diffusion
backbone serve as queries. This enables the denoising model to adapt
human--object motion to the local spatial constraints captured by the predicted
affordance.

Motion is generated autoregressively over segments. After generating each
segment, the updated human pelvis and object positions define the reference
locations for the next segment, while the corresponding goal condition
determines the goal-centered query. The three local affordance voxels are
therefore re-queried from the sequence-level affordance for each segment,
allowing the motion generator to adapt its local scene condition as the
interaction progresses.

\paragraph{Affordance-aware motion ranking.}
We use a positive SDF loss as an absolute feasibility anchor. Specifically, the 2D affordance mask ${\mathbf{A}}$ is converted into a truncated
signed distance field
$\Phi_{{\mathbf{A}}}=\mathrm{SDF}({\mathbf{A}})$,
which is normalized to $[-1,1]$ and positive inside the valid affordance region.
The anchor penalizes generated human joints that fall outside or too close to
the valid region:
\begin{equation}
\mathcal{L}_{\mathrm{sdf}}(\mathbf{x},\mathbf{A})
=
\frac{1}{T N_J}
\sum_{t=1}^{T}
\sum_{j=1}^{N_J}
\left[
\delta-
\Phi_{\mathbf{A}}
\left(
\pi(\mathbf{J}_{t,j}(\mathbf{x}))
\right)
\right]_+ ,
\end{equation}
where $\pi(\cdot)$ projects a 3D joint position onto the 2D coordinate system of the affordance map, $\Phi_{\mathbf{A}}$ is positive inside the valid affordance region and negative outside, $\delta \geq 0$ denotes a safety margin from the affordance boundary, and $[\cdot]_+=\max(\cdot,0)$. The positive anchor is
therefore
$\mathcal{L}_{\mathrm{sdf}}(\hat{\mathbf{x}}_0,{\mathbf{A}})$.
However, this positive-only objective provides limited scene-aware gradients
once most generated motions lie within feasible regions. We therefore complement
it with a hard-negative ranking objective that distinguishes alternative motion
modes.

Waypoints constrain \emph{where} an interaction should proceed but do not
uniquely determine \emph{how} the human and object should move. Even under
similar start and goal configurations, multiple plausible interaction modes may
exist within a local endpoint-conditioned motion manifold. We construct
positive-negative motion pairs $(\mathbf{x}^{+},\mathbf{x}^{-})$ with similar
relative start and goal configurations but substantially different intermediate
trajectories. Both are evaluated against
$\Phi_{\mathbf{A}^{+}}=\mathrm{SDF}(\mathbf{A}^{+})$, the SDF derived from
the positive ground-truth affordance $\mathbf{A}^{+}$. We require the
affordance-compatible positive motion to have a lower SDF penalty than the
alternative mode by at least a margin $m$:
\begin{equation}
\mathcal{L}_{\mathrm{rank}}
=
\left[
m
+
\mathcal{L}_{\mathrm{sdf}}
(\mathbf{x}^{+},\mathbf{A}^{+})
-
\mathcal{L}_{\mathrm{sdf}}
(\mathbf{x}^{-},\mathbf{A}^{+})
\right]_+ .
\end{equation}
Unlike positive-only SDF supervision, this objective remains active whenever
the two modes are insufficiently separated, providing denser relative
supervision. We define the scene-aware objective as
\begin{equation}
\mathcal{L}_{\mathrm{scene}}
=
\lambda_{\mathrm{sdf}}
\mathcal{L}_{\mathrm{sdf}}
(\hat{\mathbf{x}}_0,{\mathbf{A}})
+
\lambda_{\mathrm{rank}}
\mathcal{L}_{\mathrm{rank}} .
\end{equation}
\paragraph{Motion reconstruction and auxiliary objectives.}
In addition to the scene-aware objective, we train the motion diffusion
backbone with clean-motion and semantic-contact reconstruction losses,
together with forward-kinematics consistency for the human end effectors
and manipulated object. The complete training objective is
\begin{equation}
\mathcal{L}_{\mathrm{total}}
=
\lambda_{\mathrm{diff}}\mathcal{L}_{\mathrm{diff}}
+
\lambda_{\mathrm{fk}}\mathcal{L}_{\mathrm{fk}}
+
\mathcal{L}_{\mathrm{scene}}.
\end{equation}
We additionally apply weak classifier-free guidance following
InfBaGel~\citep{infbagel} to strengthen scene conditioning while limiting
deviation from the learned human--object motion distribution.
Detailed definitions of the auxiliary objectives are provided in the Appendix.

\paragraph{Test-time guidance.}
Following CHOIS~\citep{chois}, we retain hand--object contact and
feet--floor guidance during diffusion sampling. To separate learned scene adaptation from post-hoc geometric correction, 
we additionally evaluate a simple scene-SDF guidance that can be applied to diffusion-based motion generators without retraining:
\begin{equation}
\mathcal{G}_{\mathrm{sdf}}
=
\frac{1}{T|\mathcal{J}|}
\sum_{t=1}^{T}
\sum_{j\in\mathcal{J}}
\left[
d_{\mathrm{scene}}(\mathbf{J}_{t,j})
\right]_+ ,
\end{equation}
where $\mathcal{J}$ denotes the selected joints,
$d_{\mathrm{scene}}(\cdot)$ is positive inside occupied regions, and
$[\cdot]_+=\max(\cdot,0)$.
The guidance directly penalizes scene penetration during sampling using a small set of proxy body joints. 
We treat this guidance only as an optional test-time refinement, rather 
than as part of the proposed MAMHOI factorization. 
The complete test-time guidance formulation is provided in the Appendix.

\section{Experiment}

\subsection{Evaluation Settings}
\label{exp:evaluationset}
\paragraph{Datasets.} We train the two components of MAMHOI on complementary datasets: the scene understanding model is trained on LINGO~\citep{lingo}, while the human--object motion backbone is trained on OMOMO~\citep{omomo}. Affordance supervision for both components is constructed using relaxed axis-aligned bounding-box (AABB) regions around the corresponding motion trajectories, providing a common spatial representation across the two data sources.

For interaction-only evaluation, we use the OMOMO validation split, which contains 534 paired human--object motion sequences of 120 frames, covering 12 movable objects and 2 unseen subjects. With scene absent, this setting evaluates intrinsic interaction quality, including motion distribution, text--motion alignment, and diversity against paired ground truth.

For scene-aware evaluation, we construct a physical test set from static scenes in the LINGO test split, where paired ground-truth human--object motion is unavailable. We manually specify feasible start--goal configurations and use A* planning to generate 164 unique object paths from 20 test scenes. Evaluating five movable objects per path yields about 820 trials, which are used to assess physical feasibility such as contact and penetrations.
\paragraph{Baselines.}
We compare against CHOIS~\citep{chois}, and InfBaGel~\citep{infbagel}.
To ensure a consistent evaluation setting, all methods are initialized from the same standing human pose.
Due to the significantly different temporal horizon of the original CHOIS model, which is trained with 120-frame motion windows, we retrain CHOIS using 30-frame windows to match our setting.
We retain the native trajectory-conditioning interface of each method during evaluation.

\subsection{Evaluation Metrics}

\paragraph{Interaction-Only Metrics.}
For the OMOMO validation setting, we follow standard Text-to-HOI evaluation protocols. We report Fréchet Inception Distance (FID) to measure the distributional similarity between generated and ground-truth motions, R-precision to evaluate text-motion alignment, Multimodal Distance (MMD) to measure the distance between generated motion and text features, and Diversity to evaluate motion variation. 

\paragraph{Scene-Aware Physical Metrics.}
Following InfBaGel~\citep{infbagel}, we organize the physical metrics into three categories: \textbf{Human--Scene}, \textbf{Object--Scene}, and \textbf{Human--Object}.
For all three categories, we report $\bar{D}_{\mathrm{vtx}}$ and
$\bar{D}_{\mathrm{frm}}$ as penetration-depth metrics. For Human--Scene
penetration, we additionally report $R_{\mathrm{frm}}$, the percentage
of frames in which penetration occurs.
For Human--Scene evaluation, we report $J_{\mathrm{H}}^{\mathrm{seq}}$ to measure human-motion jerk over the sequence.
For Object--Scene evaluation, $D_{\mathrm{path}}^{\mathrm{obj}}$ measures the average distance between the generated object trajectory and the planned object path.
For Human--Object evaluation, $R_{\mathrm{con}}$ denotes the percentage of frames with active human--object contact.
Lower values indicate better performance for all metrics except $R_{\mathrm{con}}$.

\subsection{Results on OMOMO Validation}

Table~\ref{tab:omomo_validation} reports interaction-only results on the OMOMO validation split.
All methods are evaluated without any scene-specific guidance, while retaining their standard interaction-level guidance or trajectory constraints.
For our method, the affordance condition is constructed under an empty-scene assumption.
InfBaGel is evaluated with its native pelvis-trajectory constraint, which provides an explicit global motion constraint in this scene-free setting.
Therefore, these results primarily evaluate interaction and motion generation quality rather than scene-aware physical plausibility.

Compared with CHOIS, our full model improves FID, R-precision, MMD, and diversity, while InfBaGel also achieves strong retrieval performance and the highest diversity under its pelvis-constrained setting.
Interestingly, removing $\mathcal{L}_{\mathrm{rank}}$ further improves several scene-free metrics, most notably FID and MMD.
This reflects the intended role of $\mathcal{L}_{\mathrm{rank}}$ as a scene-conditioning regularizer rather than a pure motion-fidelity objective:
it discourages the generator from ignoring the scene-derived affordance condition, which can introduce a trade-off with unconstrained motion fidelity when no scene information is present.
Accordingly, its benefit is not fully captured by the scene-free OMOMO evaluation.
As shown by the scene-aware ablations in the Appendix, this regularization improves scene-condition sensitivity and reduces scene penetration.
\begin{table*}[t]
\caption{
Interaction-only evaluation on the OMOMO validation split.
All methods retain their standard non-scene guidance or trajectory constraints and no scene-specific guidance is applied.
InfBaGel is evaluated with its native pelvis-trajectory constraint.
Best results are shown in bold and second-best results are underlined.
}
\label{tab:omomo_validation}
\vspace{0.5\baselineskip}
\centering
\setlength{\tabcolsep}{7pt}
\begin{tabular}{lcccccc}
\hline
Method
& FID$\downarrow$
& \multicolumn{3}{c}{R-precision$\uparrow$}
& MMD$\downarrow$
& Diversity$\uparrow$ \\
\cline{3-5}
& & top-1 & top-2 & top-3 & & \\
\hline

CHOIS~\citep{chois}
& 2.7892
& 0.7428
& 0.8678
& 0.9062
& 3.9086
& 8.3841 \\

InfBaGel~\citep{infbagel}
& 2.2451
& \underline{0.8149}
& \underline{0.9159}
& \underline{0.9423}
& 3.4893
& \textbf{9.2597} \\

Ours
& \underline{1.8361}
& 0.7764
& \textbf{0.9279}
& \textbf{0.9543}
& \underline{3.2798}
& 8.6923 \\

Ours - $\mathcal{L}_{\mathrm{rank}}$
& \textbf{0.9403}
& \textbf{0.8173}
& 0.8750
& 0.9375
& \textbf{3.1553}
& \underline{8.7915} \\

\hline
\end{tabular}
\end{table*}

\begin{table*}[t]
\caption{
Comparison of methods on the constructed scene-aware test set.
We evaluate human--scene penetration, object--scene penetration,
object trajectory adherence, motion smoothness, and human--object
interaction quality.
$\mathcal{G}_{\mathrm{sdf}}$ denotes the same generic test-time
scene-SDF guidance applied to CHOIS and MAMHOI.
InfBaGel is evaluated with its native Bump-aware Guidance.
Best results are shown in bold and second-best results are underlined.
}
\label{tab:comparison_scene_metrics}
\vspace{0.5\baselineskip}
\centering
\setlength{\tabcolsep}{3pt}
\begin{tabular}{@{}lcccccccccc@{}}
\hline
Method
& \multicolumn{4}{c}{Human-Scene}
& \multicolumn{3}{c}{Object-Scene}
& \multicolumn{3}{c}{Human-Object} \\
\cmidrule(lr){2-5}
\cmidrule(lr){6-8}
\cmidrule(lr){9-11}

& $\bar{D}^{\mathrm{vtx}}\downarrow$
& $\bar{D}^{\mathrm{frm}}\downarrow$
& $R^{\mathrm{frm}}\downarrow$
& $J_{\mathrm{H}}^{\mathrm{seq}}\downarrow$
& $\bar{D}^{\mathrm{vtx}}\downarrow$
& $\bar{D}^{\mathrm{frm}}\downarrow$
& $D_{\mathrm{path}}^{\mathrm{obj}}\downarrow$
& $R_{\mathrm{con}}\uparrow$
& $\bar{D}^{\mathrm{vtx}}\downarrow$
& $\bar{D}^{\mathrm{frm}}\downarrow$ \\

& $\times 10^{-3}$
& $\times 10^{-3}$
& $\%$
&
& $\times 10^{-3}$
& $\times 10^{-3}$
&
& $\%$
& $\times 10^{-3}$
& $\times 10^{-3}$ \\
\hline

CHOIS
& 47.337
& 36.232
& 29.645
& 7808.0
& 0.545
& 2.267
& 0.1532
& 90.97
& \textbf{1.353}
& 2.366 \\

CHOIS + $\mathcal{G}_{\mathrm{sdf}}$
& 32.936
& 24.925
& 18.437
& 7906.9
& 0.487
& 2.041
& 0.1537
& 90.44
& 1.435
& 2.458 \\

InfBaGel
& 36.478
& 26.315
& 31.848
& 5258.7
& 3.136
& 7.281
& --
& 92.39
& 1.507
& 2.424 \\

Ours
& \underline{28.587}
& \underline{22.009}
& \underline{14.990}
& \underline{4429.3}
& \underline{0.403}
& \textbf{1.440}
& \textbf{0.1492}
& \underline{94.87}
& \underline{1.416}
& \textbf{2.200} \\

Ours + $\mathcal{G}_{\mathrm{sdf}}$
& \textbf{25.884}
& \textbf{19.889}
& \textbf{13.745}
& \textbf{4276.1}
& \textbf{0.398}
& \underline{1.498}
& \underline{0.1507}
& \textbf{95.69}
& 1.431
& \underline{2.243} \\

\hline
\end{tabular}
\end{table*}

\begin{figure}[t]
\centering
\includegraphics[width=1.02\columnwidth]{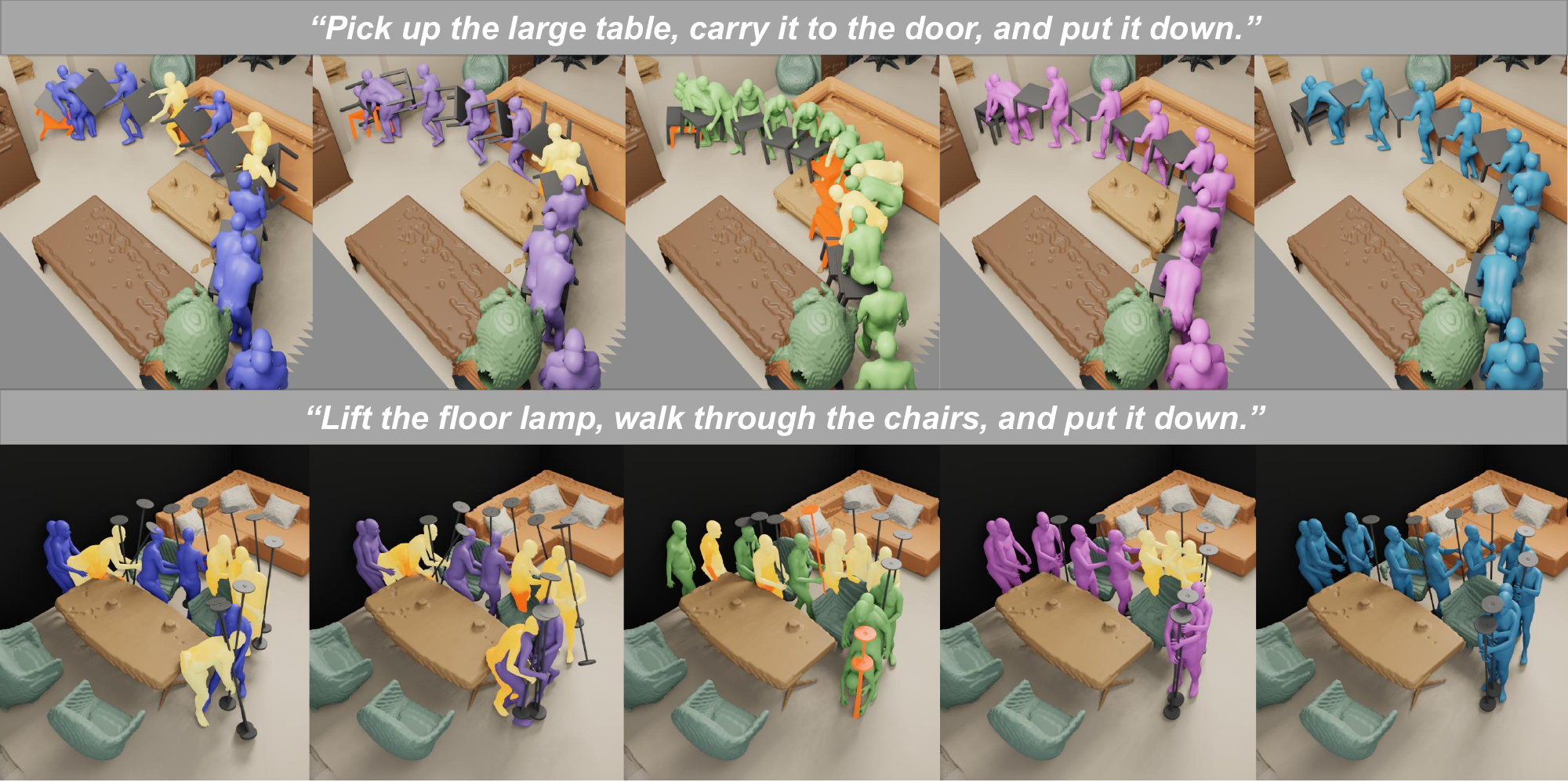}

\begin{tabular}{ccccc}
\makebox[0.18\linewidth][c]{\textbf{(a) CHOIS}} &
\makebox[0.18\linewidth][c]{\textbf{(b) CHOIS + $\mathcal{G}_{\mathrm{sdf}}$}} &
\makebox[0.18\linewidth][c]{\textbf{(c) InfBaGel}} &
\makebox[0.18\linewidth][c]{\textbf{(d) Ours}} &
\makebox[0.18\linewidth][c]{\textbf{(e) Ours + $\mathcal{G}_{\mathrm{sdf}}$}} \\
\end{tabular}

\caption{
Qualitative comparison on the constructed scene-aware test set.
$\mathcal{G}_{\mathrm{sdf}}$ denotes the generic test-time scene-SDF
refinement applied to CHOIS and MAMHOI, while InfBaGel is evaluated
with its native Bump-aware Guidance.
Penetration regions are highlighted from yellow to red.
}
\label{fig:qualitative_scene}
\end{figure}

\subsection{Results on Scene-Aware Test Set}

Table~\ref{tab:comparison_scene_metrics} and
Figure~\ref{fig:qualitative_scene} summarize the quantitative and qualitative
comparisons on the constructed LINGO-scene test set.
Even without test-time scene-SDF refinement, MAMHOI achieves lower
human--scene and object--scene penetration than CHOIS and InfBaGel,
while also improving object-path adherence and human--object contact.
Notably, the unguided MAMHOI model remains better than
CHOIS + $\mathcal{G}_{\mathrm{sdf}}$ on all reported human--scene and
object--scene measures, motion jerk, object-path error, and contact.
Qualitatively, its generated interactions more consistently follow feasible
regions in cluttered scenes while maintaining coordinated human--object
motion.
These results indicate that the scene compatibility of MAMHOI is primarily
captured by the learned affordance-conditioned motion distribution rather
than imposed through test-time geometric correction.

Applying the same $\mathcal{G}_{\mathrm{sdf}}$ refinement to CHOIS and
MAMHOI further reduces human--scene penetration for both models,
demonstrating that direct geometric correction provides a complementary
collision-reduction mechanism independent of the learned generator.
For MAMHOI, this refinement yields the lowest overall human--scene
penetration while introducing small trade-offs in frame-averaged
object--scene penetration and object-path adherence.
We do not additionally apply $\mathcal{G}_{\mathrm{sdf}}$ to InfBaGel,
as its native inference procedure already incorporates
\emph{Bump-aware Guidance}, which explicitly detects scene collisions
and iteratively pushes human and object motion away from occupied regions.
We therefore retain InfBaGel's native guidance configuration rather than
stacking an additional scene-collision objective.
Overall, the comparison separates two complementary effects:
$\mathcal{G}_{\mathrm{sdf}}$ directly suppresses penetration during sampling,
whereas MAMHOI learns a scene-adaptive interaction distribution through
affordance conditioning.

\paragraph{User Study.}
We conduct a perceptual study on the scene-aware test set following the
two-alternative forced-choice (2AFC) protocol of ZeroHSI~\citep{zerohsi}.
The study contains 136 left--right comparison pairs, where participants
select the more realistic motion generated for the same input.
We collect 884 valid votes across 52 responses and detailed results are reported in
Table~\ref{tab:user_study_full_results}.
Participants consistently prefer MAMHOI over CHOIS:
Ours is preferred over CHOIS and CHOIS+$\mathcal{G}_{\mathrm{sdf}}$
in 86.7\% and 92.3\% of votes, respectively, while
Ours+$\mathcal{G}_{\mathrm{sdf}}$ is preferred in 81.1\% and 84.3\%. 
These results provide perceptual evidence that the advantage of MAMHOI
extends beyond geometric penetration metrics.
Interestingly, $\mathcal{G}_{\mathrm{sdf}}$ does not improve human preference,
despite reducing penetration quantitatively, suggesting that post-hoc
geometric correction does not necessarily improve perceived motion naturalness.

\section{Conclusion}
We presented MAMHOI, an affordance-mediated factorization for scene-aware human-object interaction generation that separates scene-level interaction feasibility from detailed human-object motion synthesis. By using motion affordance as an explicit interface between scene reasoning and motion generation, MAMHOI allows these complementary capabilities to be learned from different sources of supervision without requiring paired human-object-scene data. Experiments in complex indoor environments show that this formulation improves scene feasibility, particularly by reducing object--scene penetration, while preserving high-quality human--object interaction dynamics, enabling more realistic and physically feasible scene-aware interactions.

\newpage
\bibliography{mamhoi}
\bibliographystyle{plainnat}

\appendix
\section{Derivation of the Factored Generation Objective}
\label{app:factored-objective}

We derive the factored objective used in Equation~\ref{eq:factored-objective}. Introducing the intermediate sequence-level affordance variable $\mathbf{A}$
and marginalizing over it gives
\begin{equation}
p(\mathbf{x}\mid c_{\mathrm{task}},c_{\mathrm{scene}})
=
\int
p(\mathbf{x}\mid \mathbf{A},c_{\mathrm{task}},c_{\mathrm{scene}})
p(\mathbf{A}\mid c_{\mathrm{task}},c_{\mathrm{scene}})
\,d\mathbf{A}.
\end{equation}
This formulation corresponds to first sampling
\begin{equation}
\mathbf{A} \sim p(\mathbf{A} \mid c_{\text{task}}, c_{\text{scene}})
\end{equation}
and then generating motion conditioned on the sampled affordance and task specification.

We assume that the affordance summarizes the scene-dependent spatial constraints required for motion generation. Once $\mathbf{A}$ is given, the remaining scene condition is therefore approximately redundant:
\begin{equation}
p(x \mid \mathbf{A}, c_{\text{task}}, c_{\text{scene}})
\approx
p(x \mid \mathbf{A}, c_{\text{task}}).
\end{equation}
We further introduce a coarse task condition $c_{\text{task}}'$ containing only the components needed for affordance prediction and assume
\begin{equation}
p(\mathbf{A} \mid c_{\text{task}}, c_{\text{scene}})
\approx
p(\mathbf{A} \mid c_{\text{task}}', c_{\text{scene}}).
\end{equation}
Substituting these two approximations into the marginalization above yields
\begin{equation}
p(x \mid c_{\text{task}}, c_{\text{scene}})
\approx
\int
p(x \mid \mathbf{A}, c_{\text{task}})
\, p(\mathbf{A} \mid c_{\text{task}}', c_{\text{scene}})
\, d\mathbf{A},
\end{equation}
which is the factored generation objective presented in Equation~\ref{eq:factored-objective}.

\section{Additional Method Details}

\subsection{Scene Understanding Model}
\subsubsection{Affordance Training Objective}
\label{app:affordance_training}

Let $\mathbf{A}^{*}\in\{0,1\}^{H\times W}$ denote the ground-truth binary motion affordance. We map it to the diffusion data space as
\begin{equation}
\mathbf{x}_0 = 2\mathbf{A}^{*}-1,
\end{equation}
and obtain the noisy affordance $\mathbf{x}_t$ using the standard forward diffusion process. Conditioned on the scene depth map, planner-provided path, and text embedding, the denoising network directly predicts the clean affordance $\hat{\mathbf{x}}_0$.

The primary denoising objective is
\begin{equation}
\mathcal{L}_{x_0}
=
\left\|
\hat{\mathbf{x}}_0-\mathbf{x}_0
\right\|_2^2.
\end{equation}

Because the task-relevant affordance typically occupies only a small portion of the scene, uniform pixel-wise supervision can under-emphasize regions relevant to the planned interaction. We therefore introduce a path-weighted binary cross-entropy loss. Let $d_i$ denote the 2D distance from pixel $i$ to the planner-provided path $\mathbf{P}$, and let $w_i$ be a distance-decaying weight with decay parameter $\tau=3$ and maximum weight $6$. The loss is
\begin{equation}
\mathcal{L}_{\mathrm{BCE}}
=
\frac{1}{HW}
\sum_{i=1}^{HW}
w_i\,
\mathrm{BCEWithLogits}
\left(
2\hat{x}_{0,i},
A_i^{*}
\right).
\end{equation}
This weighting places stronger supervision around the full object path while retaining pixel-wise supervision over the complete affordance map.

In addition, we explicitly encourage the planner-provided path to lie inside the predicted affordance. Let
\begin{equation}
\hat{\mathbf{p}}
=
\sigma(2\hat{\mathbf{x}}_0)
\end{equation}
denote the predicted affordance probabilities. For path points $\{\mathbf{p}_j\}_{j=1}^{K}$, we define
\begin{equation}
\mathcal{L}_{\mathrm{path}}
=
-\frac{1}{\sum_{j=1}^{K}\omega_j}
\sum_{j=1}^{K}
\omega_j
\log
\hat{\mathbf{p}}(\mathbf{p}_j),
\end{equation}
where the start and end points receive $\omega_j=4$, and the remaining path points use $\omega_j=1$. This term directly encourages path coverage, with additional emphasis on the initial and goal regions of the interaction.

The complete training objective is
\begin{equation}
\mathcal{L}_{\mathrm{aff}}
=
\mathcal{L}_{x_0}
+
\lambda_{\mathrm{BCE}}\,\mathcal{L}_{\mathrm{BCE}}
+
\lambda_{\mathrm{path}}\,\mathcal{L}_{\mathrm{path}},
\end{equation}
where we set
$\lambda_{\mathrm{BCE}}=0.1$
and
$\lambda_{\mathrm{path}}=0.2$
in all experiments.
\subsection{Motion Diffusion Backbone}
\label{app:motion_diffusion}

\subsubsection{Additional Training Objectives and Test-Time Guidance}
\label{app:motion_objectives}

\paragraph{Motion reconstruction and kinematic consistency.}
In addition to the scene-aware objective introduced in the main paper,
we supervise clean human--object motion reconstruction together with
semantic contact prediction:
\begin{equation}
\mathcal{L}_{\mathrm{diff}}
=
\left\|
\hat{\mathbf{x}}_0-\mathbf{x}_0
\right\|_1
+
\lambda_{\mathrm{con}}
\left\|
\hat{\mathbf{C}}-\mathbf{C}
\right\|_1,
\end{equation}
where $\hat{\mathbf{C}}$ and $\mathbf{C}$ denote the predicted and
ground-truth semantic contact states, respectively.

We further enforce forward-kinematics consistency for both the human
end effectors and the manipulated object:
\begin{equation}
\mathcal{L}_{\mathrm{fk}}
=
\left\|
\hat{\mathbf{J}}_{\mathrm{ee}}^{\mathrm{FK}}
-
\mathbf{J}_{\mathrm{ee}}
\right\|_1
+
\lambda_{\mathrm{obj}}
\left\|
\hat{\mathbf{K}}^{\mathrm{FK}}
-
\mathbf{K}
\right\|_1,
\end{equation}
where $\hat{\mathbf{J}}_{\mathrm{ee}}^{\mathrm{FK}}$ denotes the
hand and foot positions reconstructed from the predicted human joint
rotations, and $\hat{\mathbf{K}}^{\mathrm{FK}}$ denotes object
keypoints transformed using the predicted object pose.

The complete training objective is
\begin{equation}
\mathcal{L}_{\mathrm{total}}
=
\lambda_{\mathrm{diff}}\mathcal{L}_{\mathrm{diff}}
+
\lambda_{\mathrm{fk}}\mathcal{L}_{\mathrm{fk}}
+
\mathcal{L}_{\mathrm{scene}}.
\end{equation}
We use $\lambda_{\mathrm{diff}}=1$, $\lambda_{\mathrm{fk}}=0.5$,
$\lambda_{\mathrm{con}}=2$, and $\lambda_{\mathrm{obj}}=1$.
For the scene-aware objective defined in the main paper, we set
$\lambda_{\mathrm{sdf}}=0.2$, $\lambda_{\mathrm{rank}}=0.1$, and
use a ranking margin of $m=0.2$.

\paragraph{Test-time guidance.}
Following CHOIS~\citep{chois}, we retain hand--object contact and
feet--floor guidance during diffusion sampling. The hand--object
guidance encourages contacting hands to remain close to the object
surface:
\begin{equation}
\mathcal{G}_{\mathrm{con}}
=
\left\|
\mathbf{M}_{l}
\odot
\left(
\mathbf{J}^{\mathrm{hand}}_{l}
-
\mathbf{Q}^{\mathrm{obj}}_{l}
\right)
\right\|_1
+
\left\|
\mathbf{M}_{r}
\odot
\left(
\mathbf{J}^{\mathrm{hand}}_{r}
-
\mathbf{Q}^{\mathrm{obj}}_{r}
\right)
\right\|_1,
\end{equation}
where $\mathbf{M}_{l}$ and $\mathbf{M}_{r}$ are the predicted contact
masks, $\mathbf{J}^{\mathrm{hand}}_{l}$ and
$\mathbf{J}^{\mathrm{hand}}_{r}$ denote the hand positions, and
$\mathbf{Q}^{\mathrm{obj}}_{l}$ and
$\mathbf{Q}^{\mathrm{obj}}_{r}$ denote their nearest points on the
object surface.

Feet--floor guidance encourages the lower supporting foot to remain
close to the floor:
\begin{equation}
\mathcal{G}_{\mathrm{feet}}
=
\left\|
\min
\left(
(\mathbf{J}^{\mathrm{foot}}_{l})_z,
(\mathbf{J}^{\mathrm{foot}}_{r})_z
\right)
-h
\right\|_2,
\end{equation}
where $(\mathbf{J}^{\mathrm{foot}}_{l})_z$ and
$(\mathbf{J}^{\mathrm{foot}}_{r})_z$ denote the vertical positions of
the left and right feet, and $h$ is the target foot height.

For optional scene-SDF guidance, the predicted human joints are
transformed from the canonical motion frame back to the scene
coordinate system. We query the pelvis and foot joints against the
signed distance field constructed from the 2D scene occupancy map:
\begin{equation}
\mathcal{G}_{\mathrm{sdf}}
=
\frac{1}{T|\mathcal{J}|}
\sum_{t=1}^{T}
\sum_{j\in\mathcal{J}}
\left[
d_{\mathrm{scene}}
\left(
\mathbf{J}_{t,j}
\right)
\right]_{+},
\end{equation}
where $\mathcal{J}$ denotes the selected joints,
$d_{\mathrm{scene}}(\cdot)$ denotes the scene signed distance with
positive values indicating penetration, and
$[\cdot]_{+}=\max(\cdot,0)$.

The complete test-time guidance objective is
\begin{equation}
\mathcal{G}
=
\gamma_{\mathrm{con}}\mathcal{G}_{\mathrm{con}}
+
\gamma_{\mathrm{feet}}\mathcal{G}_{\mathrm{feet}}
+
\gamma_{\mathrm{sdf}}\mathcal{G}_{\mathrm{sdf}}.
\end{equation}
We use $\gamma_{\mathrm{con}}=2$ and
$\gamma_{\mathrm{feet}}=1$, and apply these two reconstruction
guidance terms only during the final 10 reverse-diffusion steps.
For the optional scene-SDF refinement, we use
$\gamma_{\mathrm{sdf}}=1.6$.
Scene-SDF guidance is disabled for the default MAMHOI model and is
enabled only for results explicitly denoted by
MAMHOI+$\mathcal{G}_{\mathrm{sdf}}$.

\section{Implementation Details}
\label{app:implementation}

\subsection{Data and Preprocessing}
\label{app:data_preprocessing}

\paragraph{Scene-aware data.}
The scene-understanding component is trained on LINGO~\citep{lingo},
which provides human motion in complex indoor environments together
with scene geometry and language descriptions.
Following the formulation in the main paper, we construct a
sequence-level motion affordance from each scene-aware motion
sequence using relaxed axis-aligned spatial support covering the
observed interaction region.
The corresponding scene geometry is represented in the horizontal
scene coordinate system, while an object-level path provides coarse
navigation structure for affordance prediction.
The scene-understanding model therefore learns to predict feasible
interaction support from the scene, planned path, and language
instruction without requiring paired human--object manipulation data.

Both the scene input and the predicted affordance map are represented
at a spatial resolution of $256\times256$.
The ground-truth binary affordance is mapped to the diffusion data
space as described in Appendix~\ref{app:affordance_training}.
At inference time, the predicted affordance probability map is
thresholded at $0.5$ before being passed to the downstream motion
generator.

\paragraph{Human--object interaction data.}
The affordance-conditioned motion generator is trained using the
processed OMOMO human--object interaction data~\citep{omomo},
following the motion representation of CHOIS~\citep{chois}.
The data contain synchronized human and object motion, text
annotations, object geometry, and semantic contact labels.
We divide each motion sequence into 30-frame windows with a stride
of 15 frames during training.
Each window is transformed into a canonical coordinate frame defined
by the initial human orientation.
Human joint positions and object center-of-mass positions are
normalized using statistics computed from the training set.

Object geometry is represented using a Basis Point Set (BPS)
descriptor containing 1,024 three-dimensional vectors.
For each training window, a reference object configuration is sampled
to construct the BPS condition and to express subsequent object
rotations relative to the reference orientation.
Windows without valid text annotations are excluded when language
conditioning is used.

\subsection{Scene Understanding Model}
\label{app:scene_understanding}

The scene-understanding model predicts the sequence-level binary
motion affordance introduced in Sec.~3.3.1 from the scene
representation, planner-provided object path, and language instruction.
Both the input scene representation and predicted affordance map have
a spatial resolution of $256\times256$.

\paragraph{Scene denoising network.}
We implement the affordance denoising network as a residual 2D U-Net.
The encoder contains three downsampling stages with channel dimensions
$64\rightarrow128\rightarrow256$, resulting in an $8\times$ spatial
downsampling at the bottleneck. Residual blocks use Group Normalization
and SiLU activations. The decoder progressively upsamples the conditioned
bottleneck features to the original spatial resolution and predicts the
clean affordance map.

\paragraph{Path and language encoding.}
The planner-provided object path is represented as a sequence of
normalized 2D coordinates. Each path point is independently encoded by
an MLP with dimensions $2\rightarrow128\rightarrow256$, followed by
Layer Normalization, producing a sequence of 256-dimensional path
features. The language instruction is encoded using a frozen CLIP
ViT-B/32 text encoder. Its 512-dimensional representation is projected
to 256 dimensions using a trainable linear projection.

\paragraph{Cross-modal conditioning.}
Task information is injected at the U-Net bottleneck through a
multi-head cross-attention layer. The spatial bottleneck features serve
as queries, while the encoded path and language features provide keys
and values. The cross-attention operates at a feature dimension of 256
with eight attention heads. The resulting conditioned spatial features
are passed to the U-Net decoder to predict the affordance map. This
design preserves the dense spatial structure of the scene while allowing
the planned path and language instruction to modulate task-relevant
regions.

\paragraph{Affordance diffusion and supervision.}
We use 1,000 diffusion timesteps and sample diffusion timesteps uniformly
during training. The denoising network directly predicts the clean
affordance $\hat{\mathbf{x}}_0$. The training objective consists of the
clean-affordance denoising loss, path-weighted binary cross-entropy loss,
and path-coverage loss defined in Appendix~\ref{app:affordance_training}.
We use $\lambda_{\mathrm{BCE}}=0.1$ and
$\lambda_{\mathrm{path}}=0.2$. The distance-decay parameter of the
path-weighted BCE is set to $3$ with a maximum weight of $6$, while the
start and goal points in the path-coverage objective receive a weight
of $4$. At inference time, the predicted probability map is thresholded
at $0.5$ to obtain the binary affordance used by the downstream motion
generator.

\paragraph{Training details.}
The scene-understanding model is optimized using Adam with a learning
rate of $1\times10^{-4}$ and a batch size of 32. We train the model for
500 epochs on a single NVIDIA A6000 GPU, requiring approximately
15 hours. The U-Net, path encoder, text projection, cross-attention
module, and affordance decoder are optimized jointly, while the CLIP
text encoder remains frozen. This component is trained independently
from the human--object motion generator.

\subsection{Affordance Interface and Motion Generator}
\label{app:affordance_motion_impl}

\paragraph{Local affordance representation.}
The predicted sequence-level affordance is lifted into the scene
coordinate system and queried locally for each motion window.
We extract three $32\times32\times32$ local volumes centered at the
current human pelvis, manipulated object, and object goal,
respectively.
Each local volume has a horizontal extent of $2.4\,\mathrm{m}$,
corresponding to the range $[-1.2,1.2]\,\mathrm{m}$ relative to its
query center and a horizontal voxel resolution of $7.5\,\mathrm{cm}$.
All three local volumes share the canonical orientation of the
current motion window.

Before feature encoding, each local volume is projected along the
vertical axis using minimum pooling, and the resulting horizontal
representation is broadcast along height.
The three processed volumes are concatenated along the channel
dimension and encoded using a ViT-based affordance encoder.
The encoder uses $8\times8$ spatial patches, an embedding dimension
of 512, six Transformer blocks, 16 attention heads, a feed-forward
dimension of 1,024, and dropout of 0.1.

The resulting 512-dimensional affordance feature conditions the
motion network through an eight-head cross-attention adapter.
The adapter additionally contains layer normalization, a
$512\rightarrow2048\rightarrow512$ feed-forward block with GELU
activation, and a learned sigmoid gate on the attention residual.
A learned null-affordance embedding is used for condition dropout and
classifier-free guidance.

\paragraph{Motion representation.}
Each frame of the diffusion state contains 220 dimensions.
The object state consists of its three-dimensional center-of-mass
position and a nine-dimensional relative rotation matrix.
Human motion is represented by the three-dimensional positions of
24 joints and continuous 6D rotations for 22 joints.
We additionally predict four semantic contact values.

The masked motion condition contains the initial human pose, initial
object pose, sparse object-motion conditions, and terminal object
position.
Object geometry is encoded from the $1024\times3$ BPS descriptor
using an MLP with dimensions $3072\rightarrow512\rightarrow256$.
Language is encoded using a frozen CLIP ViT-B/32 text encoder and
projected to a 512-dimensional feature.

\paragraph{Motion diffusion backbone.}
We use a conditional denoising diffusion model that directly predicts
the clean human--object motion $\hat{\mathbf{x}}_0$.
The noisy motion, masked motion condition, and object-geometry feature
form a 696-dimensional frame-wise input, which is projected to a
hidden dimension of 512.
The temporal backbone contains four Transformer layers with full
self-attention and four attention heads per layer.
Sinusoidal positional embeddings encode temporal position, and a
prepended conditioning token combines the projected text feature with
the diffusion-timestep embedding.

The affordance feature is injected after temporal motion encoding
through the cross-attention adapter described above.
A final linear projection predicts the complete 220-dimensional
human--object motion representation.
We use 1,000 diffusion steps with a cosine noise schedule.

During training, the affordance condition is replaced by the learned
null-affordance embedding with probability $0.2$.
This condition dropout enables classifier-free guidance over the
scene-derived affordance while leaving the remaining task conditions
unchanged.

\subsection{Motion Training Strategies}
\label{app:motion_training_strategies}

\paragraph{Orientation-balanced sampling.}
The OMOMO training data contain an imbalance in relative
human--object configurations.
We therefore apply a balanced sampling strategy based on the mean
horizontal human--object displacement within each motion window.
Windows belonging to the two selected relative-orientation groups
are sampled with inverse-frequency weighting such that they
contribute equal expected sampling mass during training.
This reduces the dominance of frequently occurring relative
interaction orientations.

\paragraph{Interaction-onset weighting.}
To emphasize the transition from approach motion to active
manipulation, we identify interaction onset using hand contact
together with object motion.
Contact is detected using a threshold of $0.5$, while object
translation and vertical displacement are detected using per-frame
thresholds of $0.01$ and $0.003$, respectively.
Events are required to persist for at least two consecutive frames.

Frames preceding the detected onset and frames within three frames
after onset receive a weight of $3$, while the remaining valid frames
receive a weight of $1$.
The weighting is applied consistently to the motion reconstruction,
semantic-contact, forward-kinematics, and object-keypoint objectives.
Each window is normalized by the sum of its valid frame weights, so
the strategy changes the relative emphasis around interaction onset
without globally increasing the contribution of an entire window.

\paragraph{Affordance-aware ranking.}
The ranking objective uses positive--negative motion pairs with
similar relative start and goal configurations but substantially
different intermediate trajectories, as described in the main paper.
Both motions are evaluated against the SDF of the positive
ground-truth affordance, and the objective encourages the positive
motion to have a lower affordance-violation cost than the alternative
motion.
We use a ranking margin of $m=0.2$ and weight this objective by
$\lambda_{\mathrm{rank}}=0.1$.
The positive SDF anchor is weighted by
$\lambda_{\mathrm{sdf}}=0.2$.

\subsection{Training and Inference}
\label{app:training_inference}

\paragraph{Independent training of the two factors.}
The scene-understanding model and the affordance-conditioned
human--object motion generator are trained independently.
The former learns the scene-to-affordance mapping from LINGO,
whereas the latter learns the affordance-to-HOI mapping from OMOMO.
Consequently, neither training stage requires paired
human--object--scene manipulation data.

\paragraph{Motion-generator training.}
The motion generator is optimized using Adam with a learning rate of
$2\times10^{-4}$ and a batch size of 128.
We train the reported model for 450,000 optimization steps on a single
NVIDIA A6000 GPU, requiring approximately 47 hours.
Diffusion timesteps are sampled uniformly during training.

The model is optimized using the clean-motion reconstruction,
semantic-contact, forward-kinematics, object-keypoint, SDF-anchor,
and affordance-ranking objectives described above.
We use $\lambda_{\mathrm{diff}}=1$,
$\lambda_{\mathrm{fk}}=0.5$,
$\lambda_{\mathrm{con}}=2$,
$\lambda_{\mathrm{obj}}=1$,
$\lambda_{\mathrm{sdf}}=0.2$, and
$\lambda_{\mathrm{rank}}=0.1$.
The ranking margin is $m=0.2$.

Gradient norms are clipped to $10$.
An exponential moving average of the model parameters is maintained
with decay $0.995$ and updated every ten optimization steps.
The EMA parameters are used for all reported motion-generation
results.

\paragraph{Autoregressive generation.}
Long-horizon interactions are generated autoregressively using
30-frame motion windows with an overlap of 10 frames, corresponding
to a stride of 20 frames.
After each window is generated, the predicted human and object states
are transformed into the canonical coordinate frame of the subsequent
window.
The local pelvis-, object-, and goal-centered affordance conditions
are then queried again from the sequence-level affordance at their
updated locations.

For overlapping windows, translations are blended linearly, while
rotations are interpolated using quaternion spherical interpolation.
This procedure allows the local affordance condition to evolve with
the generated interaction while maintaining temporal continuity
across successive windows.

\paragraph{Motion diffusion sampling.}
Inference uses ancestral reverse diffusion with 1,000 denoising
steps.
Scene classifier-free guidance combines the null-conditioned and
affordance-conditioned predictions as
\begin{equation}
\hat{\mathbf{x}}_0
=
\hat{\mathbf{x}}_0^{\mathrm{null}}
+
s
\left(
\hat{\mathbf{x}}_0^{\mathrm{aff}}
-
\hat{\mathbf{x}}_0^{\mathrm{null}}
\right),
\end{equation}
where $s$ controls the strength of affordance conditioning.
We use weak scene classifier-free guidance with
$s\in[1.0,1.3]$.
The relatively small guidance strength increases sensitivity to the
scene-derived affordance while limiting deviation from the learned
human--object interaction distribution.

Hand--object and feet--floor reconstruction guidance are applied only
during the final 10 reverse-diffusion steps, using
$\gamma_{\mathrm{con}}=2$ and $\gamma_{\mathrm{feet}}=1$.
For experiments explicitly using scene-SDF refinement, we additionally
set $\gamma_{\mathrm{sdf}}=1.6$.
The scene-SDF term is not used for the default MAMHOI results.

\section{Additional Experiments}

\subsection{Ablation of Voxel Coordinate Frame}

We study the effect of the coordinate frame used to represent local scene
geometry in Table~\ref{tab:ablation_voxel_guidance}.
Both variants use the same interaction-centric local occupancy voxels and
differ only in their orientation.
The scene-aligned variant preserves the orientation of the global scene
coordinate frame, whereas our representation rotates each voxel into the
canonical human coordinate frame.
This alignment removes variation caused by the global scene orientation and
presents nearby geometry consistently with respect to the human motion.

\begin{table*}[t]
\caption{
Ablation of the coordinate frame used for local scene voxels.
Both variants use identical interaction-centric voxel regions; the
scene-aligned variant retains the global scene orientation, while ours
canonicalizes the voxels with respect to the human orientation.
Human-aligned canonicalization consistently improves scene-aware motion
generation, with test-time SDF guidance providing additional collision
reduction.
Best and second-best results are shown in bold and underlined, respectively.
}
\label{tab:ablation_voxel_guidance}
\vspace{\baselineskip}
\centering
\setlength{\tabcolsep}{1.0pt}
\begin{tabular}{@{}lcccccccccc@{}}
\hline
Method
& \multicolumn{4}{c}{Human-Scene}
& \multicolumn{3}{c}{Object-Scene}
& \multicolumn{3}{c}{Human-Object} \\
\cmidrule(lr){2-5}
\cmidrule(lr){6-8}
\cmidrule(lr){9-11}
& $\bar{D}^{\mathrm{vtx}}\downarrow$
& $\bar{D}^{\mathrm{frm}}\downarrow$
& $R^{\mathrm{frm}}\downarrow$
& $J_{\mathrm{H}}^{\mathrm{seq}}\downarrow$
& $\bar{D}^{\mathrm{vtx}}\downarrow$
& $\bar{D}^{\mathrm{frm}}\downarrow$
& $D_{\mathrm{path}}^{\mathrm{obj}}\downarrow$
& $R_{\mathrm{con}}\uparrow$
& $\bar{D}^{\mathrm{vtx}}\downarrow$
& $\bar{D}^{\mathrm{frm}}\downarrow$ \\
& $\times 10^{-3}$
& $\times 10^{-3}$
& $\%$
&
& $\times 10^{-3}$
& $\times 10^{-3}$
&
& $\%$
& $\times 10^{-3}$
& $\times 10^{-3}$ \\
\hline
Scene-aligned
& 41.016
& 32.254
& 26.858
& 7299.8
& 0.624
& 2.185
& 0.1544
& 91.99
& 1.443
& 2.504 \\

Scene-aligned + $\mathcal{G}_{\mathrm{sdf}}$
& 32.240
& 25.043
& 17.404
& 7686.9
& 0.551
& 1.921
& 0.1535
& 93.02
& 1.487
& 2.536 \\

Ours
& \underline{28.587}
& \underline{22.009}
& \underline{14.990}
& \underline{4429.3}
& \underline{0.403}
& \textbf{1.440}
& \textbf{0.1492}
& \underline{94.87}
& \textbf{1.416}
& \textbf{2.200} \\

Ours + $\mathcal{G}_{\mathrm{sdf}}$
& \textbf{25.884}
& \textbf{19.889}
& \textbf{13.745}
& \textbf{4276.1}
& \textbf{0.398}
& \underline{1.498}
& \underline{0.1507}
& \textbf{95.69}
& \underline{1.431}
& \underline{2.243} \\
\hline
\end{tabular}
\end{table*}

Canonicalizing the local scene voxels to the human orientation consistently
improves performance over keeping them aligned with the scene coordinate
frame.
Without test-time guidance, the human-aligned representation improves all
reported metrics, with particularly large reductions in human-scene
penetration and motion jerk.
This suggests that expressing nearby geometry in a human-centric coordinate
frame provides a more consistent spatial reference for learning the mapping
between scene context and interaction motion, rather than requiring the model
to account for arbitrary global scene orientations.
Test-time SDF guidance is complementary to this representation, further
reducing human-scene penetration and improving the contact ratio.
The guided human-aligned model achieves the strongest human-scene results,
while the unguided variant retains slightly better performance on several
object-scene and human-object penetration metrics.

\subsection{Ablation of Affordance-Aware Motion Ranking}

\begin{table*}[t]
\caption{
Ablation study of the counterfactual ranking loss
$\mathcal{L}_{\mathrm{rank}}$, with and without test-time SDF
guidance. ``Ours'' denotes the model trained with
$\mathcal{L}_{\mathrm{rank}}$.
Best and second-best results are shown in bold and underlined, respectively.
}
\label{tab:ablation_ranking_loss}
\vspace{\baselineskip}
\centering
\setlength{\tabcolsep}{1.0pt}
\begin{tabular}{@{}lcccccccccc@{}}
\hline
Method
& \multicolumn{4}{c}{Human-Scene}
& \multicolumn{3}{c}{Object-Scene}
& \multicolumn{3}{c}{Human-Object} \\
\cmidrule(lr){2-5}
\cmidrule(lr){6-8}
\cmidrule(lr){9-11}
& $\bar{D}^{\mathrm{vtx}}\downarrow$
& $\bar{D}^{\mathrm{frm}}\downarrow$
& $R^{\mathrm{frm}}\downarrow$
& $J_{\mathrm{H}}^{\mathrm{seq}}\downarrow$
& $\bar{D}^{\mathrm{vtx}}\downarrow$
& $\bar{D}^{\mathrm{frm}}\downarrow$
& $D_{\mathrm{path}}^{\mathrm{obj}}\downarrow$
& $R_{\mathrm{con}}\uparrow$
& $\bar{D}^{\mathrm{vtx}}\downarrow$
& $\bar{D}^{\mathrm{frm}}\downarrow$ \\
& $\times 10^{-3}$
& $\times 10^{-3}$
& $\%$
&
& $\times 10^{-3}$
& $\times 10^{-3}$
&
& $\%$
& $\times 10^{-3}$
& $\times 10^{-3}$ \\
\hline
Ours - $\mathcal{L}_{\mathrm{rank}}$
& 30.133
& 22.890
& 17.342
& 6907.6
& 0.517
& 2.002
& \underline{0.1464}
& 92.27
& \textbf{1.394}
& 2.361 \\

Ours - $\mathcal{L}_{\mathrm{rank}}$ + $\mathcal{G}_{\mathrm{sdf}}$
& \underline{27.194}
& \underline{20.717}
& \underline{14.813}
& 6818.2
& 0.458
& 1.843
& \textbf{0.1463}
& 92.49
& \underline{1.401}
& 2.357 \\

Ours
& 28.587
& 22.009
& 14.990
& \underline{4429.3}
& \underline{0.403}
& \textbf{1.440}
& 0.1492
& \underline{94.87}
& 1.416
& \textbf{2.200} \\

Ours + $\mathcal{G}_{\mathrm{sdf}}$
& \textbf{25.884}
& \textbf{19.889}
& \textbf{13.745}
& \textbf{4276.1}
& \textbf{0.398}
& \underline{1.498}
& 0.1507
& \textbf{95.69}
& 1.431
& \underline{2.243} \\
\hline
\end{tabular}
\end{table*}

Table~\ref{tab:ablation_ranking_loss} evaluates the counterfactual ranking loss under the same local affordance representation, both with and without test-time SDF guidance. Adding \(L_{\text{rank}}\) improves scene compatibility, reducing human–scene and object–scene penetration while substantially improving motion smoothness and contact. These gains come with small trade-offs in object-path adherence and mixed human–object penetration results. These improvements indicate that the relative counterfactual signal helps distinguish motion modes that satisfy similar endpoint constraints but follow different intermediate trajectories.

Although $\mathcal{L}_{\mathrm{rank}}$ improves most scene-aware metrics,
it does not improve every metric. The variants without
$\mathcal{L}_{\mathrm{rank}}$ achieve slightly lower object-path error and
sequence-level human--object penetration, whereas adding
$\mathcal{L}_{\mathrm{rank}}$ consistently reduces human--scene and
object--scene penetration, lowers motion jerk, and increases the contact
ratio. Test-time SDF guidance is complementary and provides further
reductions in human--scene penetration.

\subsection{Results on the TRUMANS-Based Test Set}

We additionally evaluate our method on a TRUMANS-based setting to enable
comparison with HOSIG~\citep{hosig}.
Unlike our primary evaluation protocol, TRUMANS inherently contains paired
human motion and scene context.
Consequently, the motion distribution itself may already encode scene-specific
biases, introducing scene--motion distribution leakage that does not strictly
match our intended setting, where scene understanding and human--object
interaction priors are learned from separate sources.
We therefore treat this experiment as an auxiliary evaluation rather than part
of our main benchmark.
Its purpose is to examine whether the proposed factorization preserves the
information carried by both factors when trained and evaluated on a dataset
where scene and motion distributions are already correlated.

For this experiment, we train the corresponding models using the TRUMANS
data and evaluate them on the test set used by HOSIG.
The test set contains 30 planned motion sequences involving three object
categories---mouse, bottle, and cup---distributed across 20 scenes.
This evaluation is substantially smaller than our primary benchmark.
In particular, HOSIG has a comparatively expensive inference pipeline, making
large-scale evaluation impractical under the same setting.
We therefore retain the original HOSIG test split and report the results as a
controlled comparison on this common subset.

We compare against CHOIS~\citep{chois},
Lingo~\citep{lingo}, and HOSIG.
Since Lingo generates human motion but does not independently synthesize
object motion, object-related and human--object metrics are not applicable and
are omitted for this method.
HOSIG jointly models human--object interaction in scenes and therefore permits
evaluation across all three metric groups.
We note that HOSIG represents object motion using sparse keyframes followed by
post-processing.
This procedure strongly constrains the resulting interaction geometry and
leads to an almost saturated human--object contact ratio, which should be
interpreted together with its different object-motion generation mechanism.

\begin{table*}[t]
\caption{
Comparison on the TRUMANS-based scene-aware test set.
Missing or inapplicable metrics are denoted by ``--''.
Best and second-best available results are shown in bold and underlined,
respectively.
}
\label{tab:comparison_trumans_metrics}
\vspace{\baselineskip}
\centering
\setlength{\tabcolsep}{3pt}
\begin{tabular}{@{}lcccccccccc@{}}
\hline
Method
& \multicolumn{4}{c}{Human-Scene}
& \multicolumn{3}{c}{Object-Scene}
& \multicolumn{3}{c}{Human-Object} \\
\cmidrule(lr){2-5}
\cmidrule(lr){6-8}
\cmidrule(lr){9-11}
& $\bar{D}^{\mathrm{vtx}}\downarrow$
& $\bar{D}^{\mathrm{frm}}\downarrow$
& $R^{\mathrm{frm}}\downarrow$
& $J_{\mathrm{H}}^{\mathrm{seq}}\downarrow$
& $\bar{D}^{\mathrm{vtx}}\downarrow$
& $\bar{D}^{\mathrm{frm}}\downarrow$
& $D_{\mathrm{path}}^{\mathrm{obj}}\downarrow$
& $R_{\mathrm{con}}\uparrow$
& $\bar{D}^{\mathrm{vtx}}\downarrow$
& $\bar{D}^{\mathrm{frm}}\downarrow$ \\
& $\times 10^{-3}$
& $\times 10^{-3}$
& $\%$
&
& $\times 10^{-3}$
& $\times 10^{-3}$
&
& $\%$
& $\times 10^{-3}$
& $\times 10^{-3}$ \\
\hline

CHOIS
& 68.559
& 55.442
& 50.064
& 14079.9
& \underline{0.476}
& \underline{4.468}
& \underline{0.1173}
& 56.90
& \textbf{0.806}
& \textbf{2.611} \\

HOSIG
& 45.652
& 38.502
& \underline{26.572}
& 18099.7
& 7.910
& 18.200
& 0.4471
& \textbf{98.26}
& \underline{2.457}
& 4.969 \\

Lingo
& \textbf{24.541}
& \underline{19.870}
& 27.566
& \underline{13077.9}
& --
& --
& --
& --
& --
& -- \\

Ours
& \underline{25.534}
& \textbf{19.680}
& \textbf{17.282}
& \textbf{10252.5}
& \textbf{0.003}
& \textbf{0.044}
& \textbf{0.0964}
& \underline{90.11}
& 2.460
& \underline{4.694} \\
\hline
\end{tabular}
\end{table*}

As shown in Table~\ref{tab:comparison_trumans_metrics}, our method remains
competitive with models trained directly on the coupled TRUMANS
scene--motion distribution.
For human--scene penetration, it achieves the best frame-averaged penetration,
penetration rate, and motion jerk, while remaining close to Lingo in
sequence-level vertex penetration.
This is notable because Lingo is directly designed for scene-aware human motion
generation and can exploit the paired scene--motion statistics present in
TRUMANS.

Our method also achieves the lowest available object--scene penetration and
the smallest object path deviation.
These results indicate that separating scene reasoning from interaction
generation does not prevent the model from retaining effective spatial and
trajectory information when evaluated on a distribution where the two factors
are naturally correlated.
In other words, the factorized formulation does not appear to discard the
scene-related information required for motion adaptation, nor the interaction
information required for object manipulation.

For human--object interaction, HOSIG obtains the highest contact ratio.
As noted above, its object motion is generated from sparse keyframes with
subsequent post-processing, which strongly preserves contact and leads to a
near-saturated $R_{\mathrm{con}}$.
CHOIS achieves the lowest human--object penetration, whereas our method shows
higher penetration under this metric.
Taken together, this auxiliary experiment is not intended to establish the
advantages of our unpaired training formulation, since the TRUMANS setting
already contains coupled scene--motion information.
Instead, it provides complementary evidence that the proposed factorization
can preserve both scene-aware and interaction-aware information even when
applied to a distribution in which these factors are entangled.
\section{User Study Details}
\label{sec:user_study_details}

\begin{figure}[t]
\centering
\includegraphics[width=0.5\linewidth]{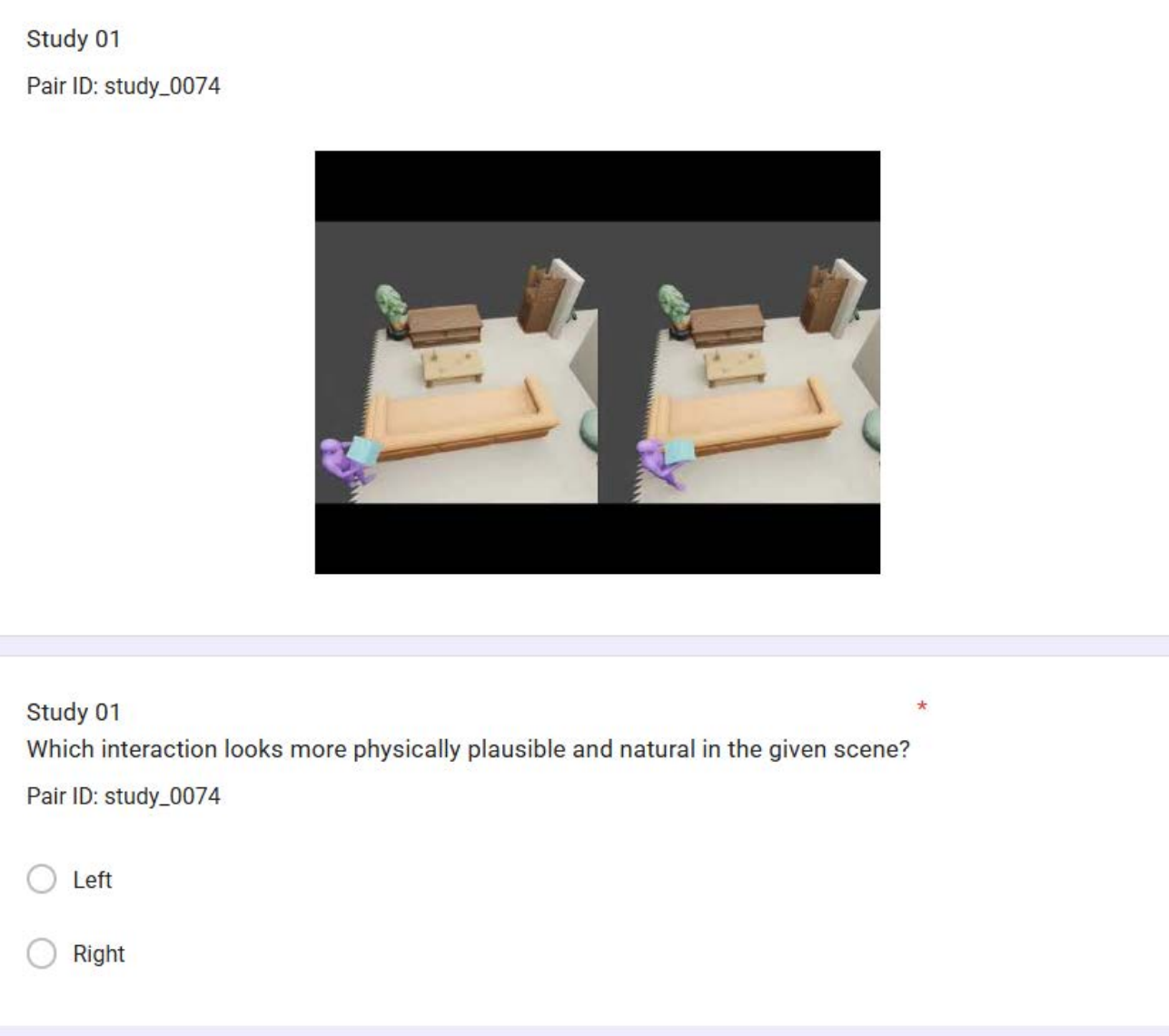}
\caption{
Example interface used in the perceptual study.
For each trial, participants view two motion sequences generated under the same scene and task configuration and select the interaction that appears more physically plausible and natural in the scene.
Method identities are hidden, while trial order and left--right placement are randomized.
}
\label{fig:user_study_interface}
\end{figure}

We conduct a two-alternative forced-choice (2AFC) perceptual study to complement the geometry-based evaluation in the main paper.
The study evaluates whether generated human--object interactions appear physically plausible and natural in complex indoor scenes.

\paragraph{Study design.}
We construct 136 pairwise comparisons from the scene-aware test set.
Each comparison presents two motion sequences generated under the same task and scene configuration.
The comparison pool contains four method pairings:
Ours vs.\ CHOIS,
Ours + SDF guidance vs.\ CHOIS + SDF guidance,
Ours vs.\ CHOIS + SDF guidance, and
Ours + SDF guidance vs.\ CHOIS.
The 136 comparisons are divided into eight tracks, each containing 17 trials, to limit the workload of each individual response.

For each trial, participants view the two generated sequences side by side and answer:
\emph{``Which interaction looks more physically plausible and natural in the given scene?''}
Participants select either the left or right sequence.
Method identities are not shown during evaluation.

\paragraph{Randomization and visualization.}
The order of trials within each track is randomized, and the left--right placement of the compared methods is independently randomized for each trial.
Both sequences in a comparison use the same rendering configuration and character colors, such that visual appearance does not reveal the underlying method.
These choices reduce potential ordering, side-selection, and rendering biases.

\paragraph{Participants and data collection.}
Approximately 48 participants were recruited through Prolific and in-person recruitment.
Some participants completed more than one track, resulting in 52 completed track-level responses in total.
Since each response contains 17 pairwise trials, we collect 884 valid pairwise judgments.
Participation was voluntary, and the collected responses are used only for research purposes and reported in aggregate.

\paragraph{Aggregation.}
For each of the four method pairings, we aggregate all binary preferences from the corresponding trials and report the percentage of judgments favoring each method.
The complete aggregated results are shown in Table~\ref{tab:user_study_full_results}.
\begin{table}[h]
    \centering
    \small
    \setlength{\tabcolsep}{4pt}
    \caption{
    Aggregated results of the perceptual study.
    ``Method A'' and ``Method B'' are used only to organize the table and do \emph{not} correspond to fixed left or right positions in the study interface; the displayed left--right placement was randomized independently for every trial.
    }
    \label{tab:user_study_full_results}
    \vspace{\baselineskip}
    \begin{tabular}{llrrrrr}
        \toprule
        \textbf{Method A} &
        \textbf{Method B} &
        \textbf{A votes} &
        \textbf{B votes} &
        \textbf{Total votes} &
        \textbf{A pref.} &
        \textbf{B pref.} \\
        \midrule
        Ours &
        CHOIS &
        195 & 30 & 225 & 86.7\% & 13.3\% \\

        Ours+$\mathcal{G}_{\mathrm{sdf}}$ &
        CHOIS+$\mathcal{G}_{\mathrm{sdf}}$ &
        182 & 34 & 216 & 84.3\% & 15.7\% \\

        Ours &
        CHOIS+$\mathcal{G}_{\mathrm{sdf}}$ &
        204 & 17 & 221 & 92.3\% & 7.7\% \\

        Ours+$\mathcal{G}_{\mathrm{sdf}}$ &
        CHOIS &
        180 & 42 & 222 & 81.1\% & 18.9\% \\
        \midrule
        \multicolumn{2}{l}{\textbf{Total judgments}} & & & \textbf{884} \\
        \bottomrule
    \end{tabular}
\end{table}

\end{document}